\documentclass{article}

\PassOptionsToPackage{numbers, compress}{natbib}
\usepackage[final]{neurips_2021}

\usepackage[utf8]{inputenc} 
\usepackage[T1]{fontenc}    
\usepackage{hyperref}       
\usepackage{url}            
\usepackage{booktabs}       
\usepackage{amsfonts}       
\usepackage{nicefrac}       
\usepackage{microtype}      
\usepackage{xcolor}         
\usepackage{graphicx}      
\title{A Conservative Q-Learning approach for handling distribution shift in sepsis treatment strategies }

\author{%
  Pramod Kaushik$^{1}$ \hspace{3mm}   Sneha Kummetha$^{2}$ \\  \hspace{-8mm}   \textbf{Perusha Moodley}$^{2}$ \hspace{3mm} \textbf{Raju S. Bapi}$^{1}$\\ \\
  $^{1}$KCIS, IIIT Hyderabad \\
  $^{2}$Independent Researcher \\
  }

\begin{document}

\maketitle

\begin{abstract}
Sepsis is a leading cause of mortality and its treatment is very expensive. Sepsis treatment is also very challenging because there is no consensus on what interventions work best and different patients respond very differently to the same treatment. Deep Reinforcement Learning methods can be used to come up with optimal policies for treatment strategies mirroring physician actions. In the healthcare scenario, the available data is mostly collected offline with no interaction with the environment, which necessitates the use of offline RL techniques. The Offline RL paradigm suffers from action distribution shifts which in turn negatively affects learning an optimal policy for the treatment. In this work, a Conservative-Q Learning (CQL) algorithm is used to mitigate this shift and its corresponding policy reaches closer to the physicians policy than conventional deep Q Learning. The policy learned could help clinicians in Intensive Care Units to make better decisions while treating septic patients and improve survival rate.
\end{abstract}

\section{Introduction}
\label{introduction}
Sepsis is a life threatening condition that is a result of body's response to severe infection and is a leading cause of mortality worldwide \cite{cohen2015sepsis}. The clinician's job of treating patients with sepsis is very challenging with a wide variety of factors to consider. Apart from controlling the source of infection and administering antibiotics, they also need to consider the dosage of vasopressers and intravenous fluids to be administered to the patients. These treatment strategies have shown drastic variations in terms of mortality rate which makes these decisions extremely critical in effectively treating a patient \cite{waechter2014interaction}.

MIMIC is a comprehensive medical dataset that contains de-identified data of critical care patients \cite{johnson2016mimic}. Deep Reinforcement Learning has been used in recent works to learn optimal policies on MIMC dataset \cite{raghu2017deep}\cite{prasad2017reinforcement}. Reinforcement Learning models are preferred over Supervised Learning models for sepsis treatment because in medical literature there is no single ground truth strategy of good treatment for sepsis. The RL frameworks developed for sepsis treatment typically follow a paradigm called Offline RL where the agent learns effective treatment strategy form previously collected static datasets without actively interacting with the environment. This is \citet{raghu2017deep} developed one such offline Deep Reinforcement learning model to identify Sepsis patients admitted in ICU so that Vasopressor and Intravenous fluid interventions can be done at the early stages of treatment to improve their chances of survival. They used a discrete action space, continuous state-space modelling using patient's physiological data from the ICU as a continuous vector and a Deep Q-Network to find suitable actions (Vasopressor and IV fluid dosages). However, one of the limitations of this work was that, the model didn't perform very well when the patient is very critical (high SOFA score). In this work, we address this limitation by using a method called Conservative Q-Learning which is capable of handling distributional shifts which is an issue commonly observed in offline RL methods.

\section{Preliminaries}
\label{prelims}

In the Reinforcement Learning framework, the agent tries to maximise its total reward by taking actions in an environment. The actions taken by the agent influence the next state and the reward it receives from the environment. The MDP formalism is used to model this behaviour which is represented as a 5-tuple 
\begin{math}\langle\mathcal{S}, \mathcal{A}, r, p, \gamma\rangle \end{math}
where at a timestep \textit{t}, the agent is in state 
\begin{math}s_{t} \in \mathcal{S}\end{math}, 
and it takes action 
\begin{math}a_{t} \in \mathcal{A}\end{math},
that leads to the next state 
\begin{math}s_{t+1} \in \mathcal{S}\end{math} according to the transition probability 
\begin{math}p\left(s^{\prime} \mid s, a\right)\end{math}.

 The agent maximises the expected discounted reward and is defined as \begin{math}R_{t}=\sum_{t^{\prime}=t}^{T} \gamma^{t^{\prime}-t}  r_{t^{\prime}}\end{math} where $\gamma$ is the discount factor and it denotes the tradeoff between immediate and future rewards, and T is the terminal timestep. The
optimal action value function, \begin{math}Q^{*}(s, a)=\max _{\pi} \mathbb{E}\left[R_{t} \mid s_{t}=s, a_{t}=a, \pi\right]\end{math}, is the maximum discounted
expected reward obtained after following policy $\pi$ which is a mapping from states to actions.

Here in the offline RL  (or batch RL) setting, it is  required that the policy is learnt solely from a previously collected dataset, without any active interactions with the environment. Offline reinforcement learning scenarios create issues for existing off-policy algorithms like Deep Q learning even though these methods can learn from off-policy data. These class of algorithms cannot fully learn offline without additional on-policy data. Deep neural networks which are highly expressive and high dimensional worsen the issue since function approximation leaves algorithms susceptible to distribution shift and creates overestimation of values in these standard off-policy algorithms. One of the more effective approaches to counter this scenario is by regularising the value function or Q function directly to avoid overestimation for out-of-distribution actions \cite{kumar2020conservative}. This over estimation by neural networks happens for value of actions which are not frequently observed or unobserved in the data like in the case of patients with high SOFA scores.

An effective approach in the literature of offline RL to counter overestimation is by regularizing the value function or Q-function directly to avoid overestimation for out-of-distribution actions \cite{kumar2020conservative}. The loss function with the regulariser is given by the equation:
\begin{equation}
\label{CQL_equation1}
L(\theta) = \alpha \mathbb{E}_{s_t \sim D}
            [\log{\sum_a \exp{Q_{\theta}(s_t, a)}}
             - \mathbb{E}_{a \sim D} [Q_{\theta}(s, a)]]
            + L_{DoubleDQN}(\theta)
\end{equation}

This conservative penalty can be interpreted in a intuitive manner: the log-sum-exp would be heavily influenced by the action with the largest Q-value, and therefore this type of penalty tends to minimize whichever Q-value is largest at each state. The second term ensures maximising the values for state-action values in the batch. Intuitively, this acts to ensure that high Q-values are only assigned to in-distribution action.

\section{Methods}
\subsection{Data and preprocessing}
\label{preprocessing}
We obtained the data from Multiparameter Intelligent Monitoring in Intensive Care (MIMIC-III v1.4) \cite{moody1996database}\cite{johnson2016mimic} database from Physionet \cite{goldberger2000physiobank} and focused only on patients who fulfilled the Sepsis-3 criteria \cite{singer2016third}. We setup a local instance of the MIMIC database and extracted the relevant physiological parameters of each patient (like age, gender, vital signs etc) and aggregated the data points into 4 hour intervals as described in \citet{raghu2017deep}. We performed the detailed pre-processing steps mentioned in \citet{komorowski2018artificial} and the code for the pre-processing steps is located here : \url{https://github.com/matthieukomorowski/AI_Clinician}. All this pre-processing to get the required features yielded 48 features that denote each state. 

\subsection{Actions and rewards}
\label{actions}
The actions are defined in a discrete 5 X 5 action space which represent the interventions spanning the space of intravenous fluids(IV) and maximum Vasopresser (VP) dosage in 4 hour window. All non-zero dosages of both the drugs were divided into 4 quartiles and each quartile was represented as an integer (1-4) representing its quartile bin for each drug. The action space includes a special case of 0 bin which represents no drug given. The actions were then represented as tuple (total IV, max VP) at each timestep. 

Since the objective is to take actions that will improve the patient's condition, the reward function should be clinically guided such that the model penalizes those actions that deteriorate the patient's condition and rewards the actions that improve their condition. The indicators of patient health used to develop the reward function are SOFA (organ failure measure) score and Lactate levels (septic patients have high levels) of the patient. At the terminal time steps, the patients who survived the hospital stay are issued a reward of +15 while those who passed away are given a reward of -15 as described in \citet{raghu2017deep}.

\subsection{Model architecture}
\label{architecture}
Inorder to compare our performance to that of \citet{raghu2017deep}, we used the same underlying network and configuration which is a fully-connected Dueling Double-Deep Q Network with two hidden layers of size 128. To the existing Network, we applied CQL as explained in Section \ref{prelims}. We used alpha = 0.1 in Equation \ref{CQL_equation1}

Implementation of reward functions, action space discretization and the base model that is used from \citet{raghu2017deep} can be found here : \url{https://github.com/aniruddhraghu/sepsisrl}. The offline RL CQL implementation that is presented in the current work can be found here : \url{https://github.com/xercerfe/sepsisCQL}

\section{Results}
To understand how the model performs in various severity groups, we performed qualitative analysis. It was divided into 3 groups: the first group had SOFA scores less than 5 (low SOFA group), the second group had SOFA scores between 5 to 15 (medium SOFA group) and the third group had SOFA scores greater than 15 (high SOFA group).

\begin{figure}
  \centering
  \includegraphics[width=\linewidth]{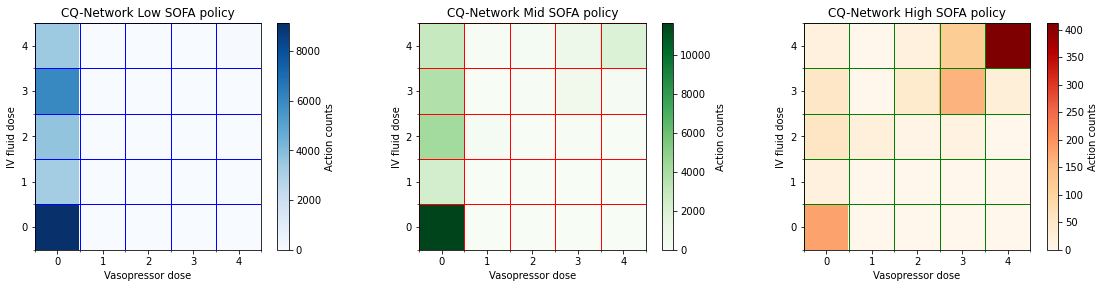}
  \includegraphics[width=\linewidth]{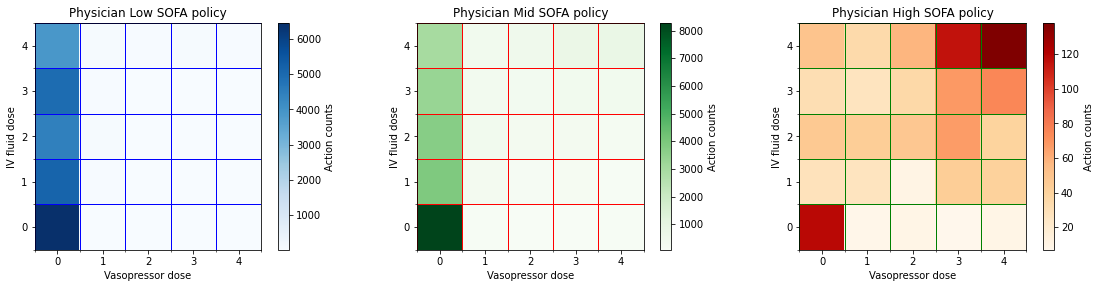}
  \includegraphics[width=\linewidth]{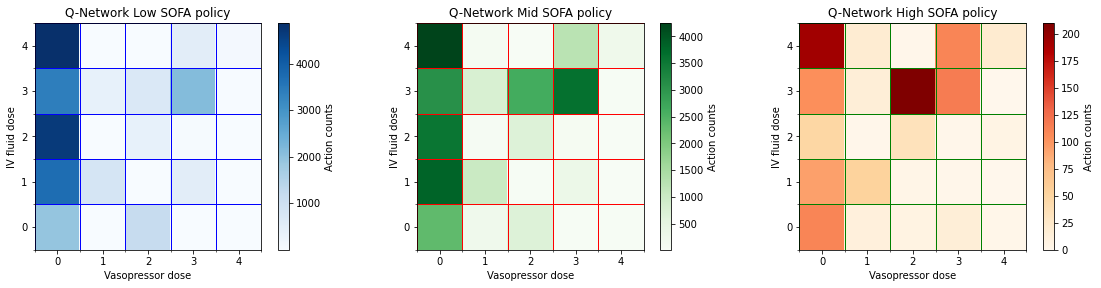}
  \caption{Policies learned by the different models, as a 2D histogram, where all actions selected by the physician and model are aggregated on the test set over all the severity groups. The axis labels are the discretised action space, where 0 represents no drug given, and 4 the maximum of that particular drug. The Q Network learns to prescribe low dosage of Vasopressers in low and medium SOFA scenarios while the CQ Network model learns to prescribe low dosage of vasopressors in low and medium SOFA groups and high dosage in the high SOFA group which is a key feature of the physician’s policy. }
  \label{2d_hist}
\end{figure}

Figure \ref{2d_hist} illustrates the treatment policies followed by the model and the physicians for the three different severity groups. The action numbers on both the axes represent the different discrete actions selected at any given timestep, and each chart shown is an aggregate of actions taken over all timesteps for those severity groups. The action zero indicates that no drugs were given to the patient at that timestep and the higher actions represent higher drug dosages. Both the drug dosages are divided into quartiles and each each action represents a quartile.

We plotted 3 histograms in Figure \ref{2d_hist}. The first row (CQ Network) is our model, the second row is the Physician model and the third row is the Q Network model reproduced from \citet{raghu2017deep}. As seen in Figure \ref{2d_hist}, the physicians hardly ever prescribe Vasopressors unless SOFA score is high (the density of actions is higher for action 0 of Vasopressors in low and medium SOFA) which is reflected in the physician policy. This is clinically accurate because Physicians in general don't tend to prescribe high doses of Vasopressors to Septic patients as majority of them are not hypotensive (Vasopressers are administered when the patient is hypotensive). But when the patient is very critical, which is the case in high SOFA scenario, physicians administer high doses of Vasopressers which is seen in the Physician policy as well. As shown in the figures, CQ Network's behaviour is very similar to what is observed in the Physician policy in all the categories. The CQ Network approximates Physician policy better than the Q Network in the high SOFA case.

\begin{figure}
  \centering
  \includegraphics[width=\linewidth, height=0.4\linewidth]{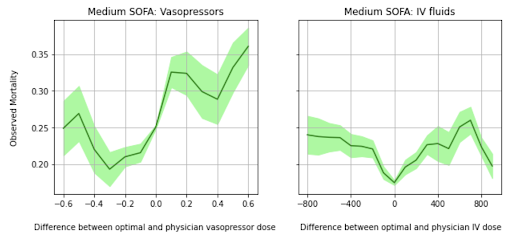}
  \includegraphics[width=\linewidth, height=0.4\linewidth]{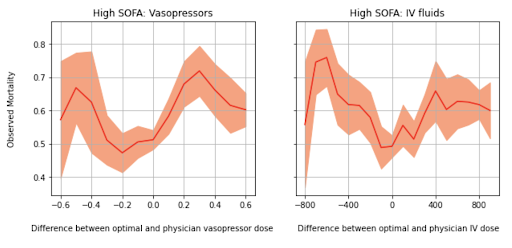}
  \caption{Relationship between mortality (y axis) and the difference in dosage suggested by the CQ policy and the Physician policy (x axis). Low mortality is observed in both medium and high SOFA groups when the difference between the optimal policy dosage and clinician dosage is low}
  \label{mortality_rates}
\end{figure}

Figure \ref{mortality_rates} shows correlation between mortality and the difference between the dosages suggested by the policy and the dosages suggested by physician policy for both the interventions. We have only plotted medium and high SOFA groups because the mortality rate is very low for low SOFA scenario and hence not very interesting as explained in \citet{raghu2017deep}. We can see that when the difference between dosage suggested by the policy coincides with the dosage suggested by physician policy (when x is 0), mortality rate is lowest for both interventions. When the difference increases, mortality rate also increases. This is observed in both medium and high SOFA groups suggesting that the model performs well in both cases, which is similar to how a clinician prescribes the interventions leading to lower mortality rates. 

\section{Conclusions}
\label{conclusions}
In this work, we applied Conservative-Q Learning to the existing Dueling Double-Deep Q Networks to address distributional shift caused by fewer data points present in the high SOFA scenario. Through qualitative analysis, we demonstrated that CQL helps learning optimal policy for patient treatment even when they're in critical condition (high SOFA). However, quantitative analysis needs to be performed using an Off policy evaluation method to determine the robustness of this model, if it were to be deployed. This is an important future direction for this work.

\section{Broader Impact}
\label{impact}
The CQ Network model in this work was evaluated qualitatively but after performing quantitative analysis with a robust off policy evaluation method, this model could be deployed in a hospital to help physicians during medical decision making. Over-reliance on the model for critical decisions could be a potential negative impact especially since these decisions can have very fatal consequences on patients being treated.

\begin{ack}
Use unnumbered first level headings for the acknowledgments. All acknowledgments
go at the end of the paper before the list of references. Moreover, you are required to declare
funding (financial activities supporting the submitted work) and competing interests (related financial activities outside the submitted work).
More information about this disclosure can be found at: \url{https://neurips.cc/Conferences/2021/PaperInformation/FundingDisclosure}.

Do {\bf not} include this section in the anonymized submission, only in the final paper. You can use the \texttt{ack} environment provided in the style file to autmoatically hide this section in the anonymized submission.
\end{ack}

\bibliography{bibFile}

\section*{Checklist}

\begin{enumerate}

\item For all authors...
\begin{enumerate}
  \item Do the main claims made in the abstract and introduction accurately reflect the paper's contributions and scope?
    \answerYes{} {\bf Yes, we claimed to address distributional shift using CQL and that's what we did}
  \item Did you describe the limitations of your work?
    \answerYes{} {\bf See Section \ref{conclusions}}
  \item Did you discuss any potential negative societal impacts of your work?
    \answerYes{} {\bf See Section \ref{impact}}
  \item Have you read the ethics review guidelines and ensured that your paper conforms to them?
    \answerYes{} Yes, we have read the ethics review guidelines and our paper conforms to them.
\end{enumerate}

\item If you are including theoretical results...
\begin{enumerate}
  \item Did you state the full set of assumptions of all theoretical results?
    \answerNA{} {\bf This is not a theoretical paper}
	\item Did you include complete proofs of all theoretical results?
    \answerNA{} {\bf This is not a theoretical paper}
\end{enumerate}

\item If you ran experiments...
\begin{enumerate}
  \item Did you include the code, data, and instructions needed to reproduce the main experimental results (either in the supplemental material or as a URL)?
    \answerYes{} {\bf See Section \ref{preprocessing} for pre-processing instructions and Section \ref{architecture} for all the URLs inorder to replicate the experimental results}
  \item Did you specify all the training details (e.g., data splits, hyperparameters, how they were chosen)?
    \answerYes{} {\bf See Section \ref{actions}, Section \ref{architecture} and Appendix \ref{features} }
	\item Did you report error bars (e.g., with respect to the random seed after running experiments multiple times)?
    \answerYes{} {\bf See Figure \ref{mortality_rates}}
	\item Did you include the total amount of compute and the type of resources used (e.g., type of GPUs, internal cluster, or cloud provider)?
    \answerNA{} {\bf We didn't use GPUs}
\end{enumerate}

\item If you are using existing assets (e.g., code, data, models) or curating/releasing new assets...
\begin{enumerate}
  \item If your work uses existing assets, did you cite the creators?
    \answerYes{} {\bf Mentioned the code used for pre-processing in Section \ref{preprocessing} and for main architecture used in Section \ref{architecture} }
  \item Did you mention the license of the assets?
    \answerNA{} {\bf None of the assets we used require a license}
  \item Did you include any new assets either in the supplemental material or as a URL?
    \answerYes{} {\bf See Section \ref{preprocessing} and Section \ref{architecture}}
  \item Did you discuss whether and how consent was obtained from people whose data you're using/curating?
    \answerNA{} {\bf MIMIC dataset that we are using is publicly available}
  \item Did you discuss whether the data you are using/curating contains personally identifiable information or offensive content?
    \answerYes{} {\bf Explained in Section \ref{introduction} that the data is deidentified patient data}
\end{enumerate}

\item If you used crowdsourcing or conducted research with human subjects...
\begin{enumerate}
  \item Did you include the full text of instructions given to participants and screenshots, if applicable?
    \answerNA{} {\bf We did not conduct research on human subjects} 
  \item Did you describe any potential participant risks, with links to Institutional Review Board (IRB) approvals, if applicable?
    \answerNA{} {\bf We did not conduct research on human subjects}
  \item Did you include the estimated hourly wage paid to participants and the total amount spent on participant compensation? 
    \answerNA{} {\bf We did not conduct research on human subjects}
\end{enumerate}

\end{enumerate}


\appendix

\section{Appendix}
\subsection{State features used in the model}
\label{features}
We followed the same approach as \citet{raghu2017deep} to pick the state features that best represent what a clinicians would consider while deciding the best treatment strategy for septic patients. The complete list of physiological features used are :

Age, Gender, Shock Index, Readmission, Elixhauser, Glasgow Coma Scale (GCS), SIRS, Sequential Organ Failure Assessment (SOFA), Arterial Lactate, Bicarbonate, International Normalized Ratio (INR), Sodium,  White Blood Cell Count, CO2, Creatinine, Ionised Calcium, Serum Glutamic-Oxaloacetic Transaminase (SGOT), Prothrombin Time (PT), Platelets, Count, Total bilirubin, Albumin, Calcium, Glucose, Hemoglobin, Partial Thromboplastin Time (PTT), Potassium, Serum Glutamic-Pyruvic Transaminase (SGPT), Arterial Blood Gas, BUN - Blood Urea Nitrogen, Chloride, Arterial pH, Magnesium, Diastolic Blood Pressure, Mean Blood Pressure, Respiratory Rate, SpO2, Systolic Blood Pressure,  PaCO2, PaO2, FiO2, PaO/FiO2 ratio, Temperature (Celsius), Weight (kg), Heart Rate, Total Fluid Output, Mechanical Ventilation, Fluid Output - 4 hourly period.

\end{document}